\documentclass[conference]{IEEEtran}
\usepackage{times}
\usepackage{graphicx}
\usepackage{xcolor}
\usepackage{capt-of}
\usepackage{caption}

\usepackage[numbers]{natbib}
\usepackage{multicol}
\usepackage[bookmarks=true]{hyperref}

\usepackage{booktabs}
\usepackage{array}
\usepackage{multirow}
\usepackage{amsmath}
\usepackage{amssymb}
\usepackage{algorithm}
\usepackage{algpseudocode}
\usepackage{placeins}

\usepackage{xcolor}

\definecolor{metabg}{RGB}{242,242,242}

\hypersetup{
  pdftitle={Tau0-VLA: A Hierarchical VLA Foundation Model with World-Model-Guided Test-Time Computation},
  pdfauthor={Xiaowei Cai et al.},
  hypertexnames=false,
  hidelinks
}

\newcommand{\myparagraph}[1]{\par\noindent\textbf{#1}\quad}

\renewcommand{\footnoterule}{%
  \kern-3pt
  \hrule width 0.4\columnwidth
  \kern2.6pt
}

\begin{document}

\title{$\tau_0$-VLA: a Hierarchical Robot Foundation Model with World-Model-Guided Test-Time Computation }

\author{
\IEEEauthorblockN{\small
Xiaowei Cai\textsuperscript{2},
Yunuo Cai\textsuperscript{1,2},
Bingao Chen\textsuperscript{2,3},
Jingxiao Chen\textsuperscript{2},
Zhi Chen\textsuperscript{2},
Siyuan Feng\textsuperscript{2},
Tengyu Hou\textsuperscript{2},
Jingshun Huang\textsuperscript{1,2},\\
Han Jiang\textsuperscript{2},
Runkun Ju\textsuperscript{2},
Dong Li\textsuperscript{2},
Mingxiang Li\textsuperscript{2},
Shaowei Li\textsuperscript{2},
Xinchen Li\textsuperscript{2},
Yifan Li\textsuperscript{1,2},
Yi Liu\textsuperscript{1,2},\\
Zhongyuan Liu\textsuperscript{2},
Jianlan Luo\textsuperscript{1,2},
Junwen Miao\textsuperscript{2},
Ruiqi Ni\textsuperscript{2},
Buqing Nie\textsuperscript{2},
Mingjie Pan\textsuperscript{1,2},
Xinlin Ren\textsuperscript{2},
Jianheng Song\textsuperscript{2},\\
Jiaxu Wang\textsuperscript{2,3},
Peiqi Wang\textsuperscript{2},
Sen Wang\textsuperscript{2},
Xiaoyan Wang\textsuperscript{2},
Dafeng Wei\textsuperscript{2},
Dongming Wu\textsuperscript{2,3},
Pengwei Xie\textsuperscript{2},
Pu Yang\textsuperscript{2},\\
Hangjian Ye\textsuperscript{1,2},
Xiangyu Yue\textsuperscript{2,3},
Jinyu Zhang\textsuperscript{1,2},
Qinglin Zhang\textsuperscript{2},
Xueyong Zhao\textsuperscript{2},
Pengfei Zhou\textsuperscript{2},
Yue Zhou\textsuperscript{2}}
\vspace{0.6em}
\IEEEauthorblockA{
\textsuperscript{1}Shanghai Innovation Institute \qquad
\textsuperscript{2}Agibot Finch \qquad
\textsuperscript{3}The Chinese University of Hong Kong\\[0.4em]
\href{https://tau0-vla.github.io/}{\textcolor[HTML]{BE0027}{https://tau0-vla.github.io/}}}
\thanks{Authors are listed alphabetically. See the Author Contributions section for individual contributions.}
}

\IEEEoverridecommandlockouts
\IEEEaftertitletext{%
  \vspace{-2.5em}
  \begin{center}
    \includegraphics[width=0.98\textwidth]{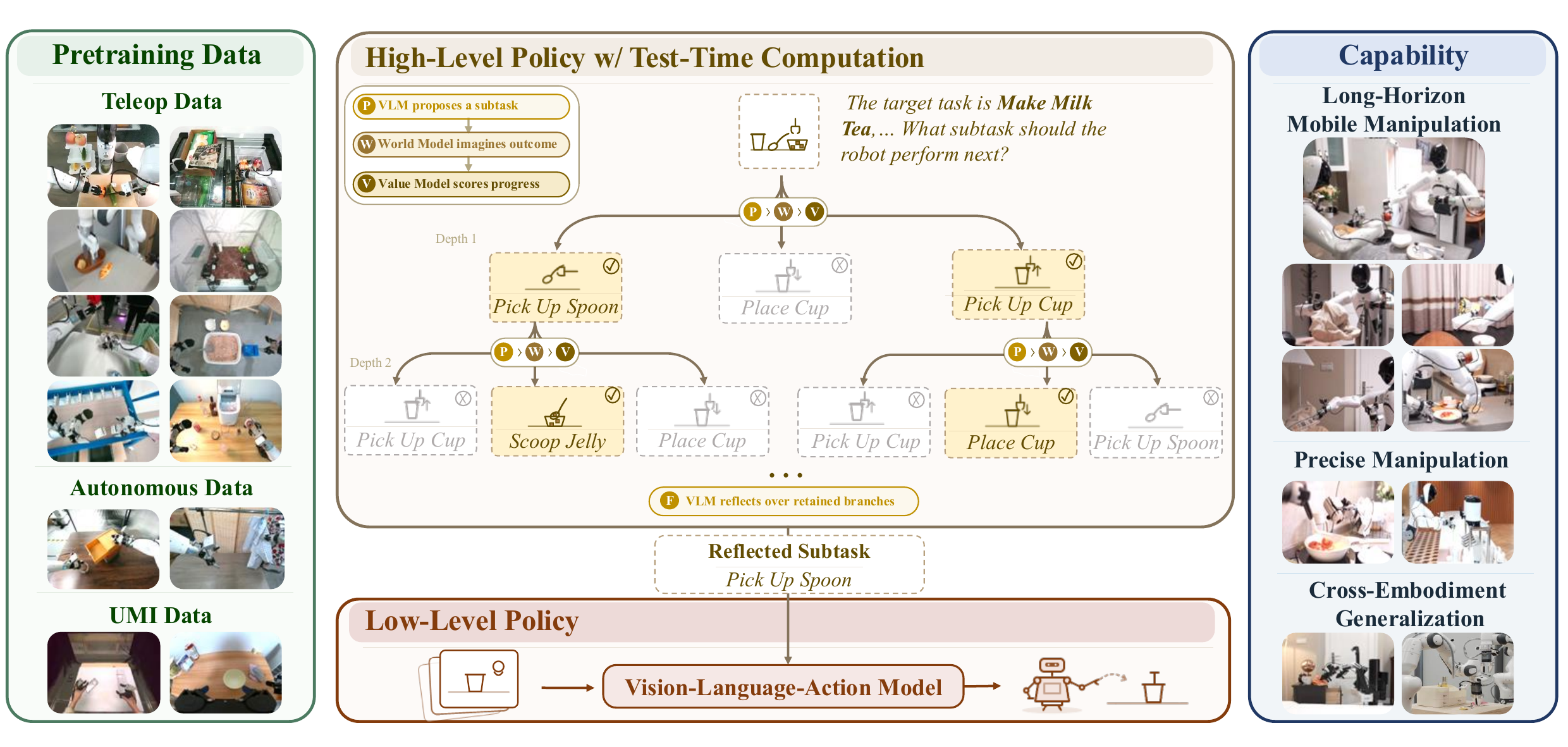}
\captionof{figure}{\textbf{Overview of $\tau_0$-VLA.}
The high-level policy uses world-model-guided test-time computation to search over subtask sequences. At each expansion step, a VLM proposes candidate subtasks, a world model predicts their visual outcomes, and a value model evaluates the resulting task progress. Beam search retains promising branches, after which a reflective model commits to the next subtask based on the retained candidates and their predicted consequences. The figure illustrates a single candidate expansion with $N=3$; recursive beam expansion to depth $D$ is omitted for brevity. The selected subtask then conditions a low-level VLA policy trained on $40{,}115$ hours of heterogeneous real-world data, enabling long-horizon mobile manipulation, precise execution, and deployment across multiple robot embodiments.}
\label{fig:overview}

  \end{center}
  \vspace{0.3em}
}



%

\maketitle

\begin{abstract}
Long-horizon robot manipulation requires a robot to both execute individual skills reliably and sequence them coherently over extended tasks. Most hierarchical vision-language-action (VLA) models make each such decision with a single forward pass, leaving no mechanism to allocate additional computation to difficult or consequential choices.
We introduce $\tau_0$-VLA, a hierarchical robot foundation model that formulates high-level subtask generation as a compute-scalable inference problem through world-model-guided test-time computation.
At each inference step, the high-level policy uses execution memory to generate a subtask and, when needed, searches over alternatives before committing to its output.
A low-level policy then executes the generated subtask across multiple robot embodiments. The policy is trained on 40,115 hours of heterogeneous real-world data with multimodal co-training. Across in-domain and distribution-shifted settings, allocating additional test-time computation substantially improves next-subtask prediction accuracy, and these gains translate into higher closed-loop success on long-horizon robot manipulation tasks.
\end{abstract}


\section{Introduction}
Long-horizon manipulation encompasses a broad class of real-world robotic
tasks, such as cleaning a room, cooking a meal, or preparing a drink. These
tasks require coherent sequences of subtasks carried out over minutes or hours.
The robot must locate and manipulate objects, interact with articulated
structures, verify intermediate outcomes, and recover from local failures.
Such tasks are not simply long sequences of motor commands; they are
sequences of consequential decisions.
Throughout execution, the robot must infer what has been completed, determine
what remains, and maintain an appropriate subtask command.
An incorrect subtask choice cannot generally be corrected by more precise
motor control: the robot may execute the wrong subtask perfectly.

Language-conditioned robot policies and vision-language-action (VLA) models provide a natural foundation for long-horizon robotic control by mapping visual observations and language instructions to robot actions~\citep{rt1,rt2,openvla,octo,pi0,pi05}. Hierarchical robot policies further connect semantic reasoning to visuomotor control through language commands, explicit subtask interfaces, or latent representations~\citep{saycan,rth,hirobot,groot,helix,gemini_robotics,pi05,pi07}. Many hierarchical systems still make each high-level decision with a fixed inference budget, directly mapping the current observation and execution history to the next subtask. In this setting, the model neither compares alternative subtasks nor evaluates them through the physical states they are expected to produce. Recent systems instead use high-level tree search or recursive subgoal refinement~\citep{vine,seeing_farther,anticipation_vla}, motivating the question we study here: how to couple open-ended language-subtask search with visual outcome prediction and execution memory in a generalist real-robot hierarchy.
Without pre-commitment evaluation, an incorrect decision is typically detected only after execution has already altered the environment, at which point replanning can respond but cannot recover the cost of the failed commitment. Thus, the central bottleneck in long-horizon execution is no longer the availability of a hierarchical action interface, but the inference procedure used to generate the next subtask.

We instead formulate generating the next subtask as an inference-time reasoning problem, allowing computation to scale with the difficulty of each decision, as test-time computation does in language models~\citep{o1,deepseek_r1}. The subtask is a natural unit for this reasoning. Action-level search can resolve local control ambiguities, but often exposes only local physical consequences~\citep{vgps,robomonkey,rover,forewarn,vla_reasoner,wla0}. Language-only reasoning operates over longer horizons, but compares alternatives without grounding them in the physical states they would produce. Subtasks lie between these extremes: they are sparse enough to justify additional computation, yet temporally extended enough to induce meaningful and distinguishable changes in the environment. Comparing candidate subtasks before execution therefore requires predicting the state that each candidate would produce. Generative models of future observations provide precisely this capability~\citep{unipi,susie,vlp,reflective_planning,vista}, enabling candidate subtasks to be evaluated through their anticipated physical consequences before they are issued for execution.

We present $\tau_0$-VLA, a hierarchical VLA system that instantiates this approach. At each inference step, a memory-augmented high-level policy uses the current observation and execution memory to generate the appropriate subtask. The proposal is used directly when the policy is confident. Otherwise, token-confidence statistics trigger an additional reasoning procedure. This procedure follows a propose--predict--evaluate loop. The proposal model generates open-ended candidate subtasks, a world model predicts the terminal observation induced by each candidate, and a value model scores candidate quality from the predicted outcomes. Beam search recursively expands promising candidates, allowing the inference budget to scale through the branching factor, beam width, and search depth. A reflective model then conditions on the retained branches and generates the final subtask. Its output may coincide with a proposal but is not restricted to the candidate set. The low-level policy then executes the generated subtask. The resulting real observation is incorporated into execution memory at the next inference step, closing the loop between predicted and observed consequences. To support reliable reasoning over extended executions, we train the memory mechanism to correct its own state. Specifically, we perturb memories derived from existing demonstrations and train the high-level policy to repair records that lag behind, run ahead of, or otherwise misrepresent the robot's actual progress, requiring no additional annotation.

Once generated, each subtask is executed by a generalist low-level policy that combines a pretrained vision-language backbone with a Mixture-of-Transformers action expert. The policy operates over a unified $40$-dimensional state and action space covering end effectors, arm joints, grippers, the waist, and the mobile base. This unified representation allows a single model to support fixed-base manipulation, bimanual coordination, and mobile whole-body control across multiple robot embodiments. We train the policy on approximately $40{,}000$ hours of robot data collected from heterogeneous sources, together with multimodal co-training data, yielding a language-conditioned execution interface without the need for subtask-specific controllers.

We evaluate the complete system across multiple robot embodiments on real-world,
long-horizon manipulation tasks. The evaluation covers room cleaning, meal
preparation, tea making, and laundry collection, with episodes lasting up to
12 minutes.
Hierarchical test-time computation substantially improves task success over whole-task inference using the same low-level policy and increases next-subtask prediction accuracy, with further gains obtained from larger inference budgets.

Our central contribution is a complete VLA system, $\tau_0$-VLA, that formulates high-level subtask generation as a compute-scalable inference problem while retaining a unified low-level control interface across
multiple robot embodiments.
Our experiments show that allocating additional computation to high-level
decisions substantially improves next-subtask prediction, and yields higher end-to-end task success on
long-horizon real-robot tasks.
We further provide a detailed empirical analysis of execution memory, consequence-aware search, and the compute-accuracy trade-off.

\section{Related Work}

$\tau_0$-VLA brings together three lines of work: generalist VLA models
for low-level control, hierarchical robot policies for subtask-level
decision making, and world models with test-time computation for evaluating
candidate decisions. We thus survey works in these areas.
\subsection{Vision-Language-Action Models}

Vision-language-action (VLA) models map visual observations and language
instructions to robot actions, often by adapting pretrained vision-language
backbones for continuous control. Early work established scalable
language-conditioned robot learning and transfer from vision-language
pretraining~\citep{rt1,rt2}. Subsequent work broadened this paradigm across
tasks and embodiments~\citep{openvla,octo}, while recent generalist models
target continuous control, open-world generalization, and cross-embodiment
transfer~\citep{pi0,pi05,groot,xvla,lingbot,gemini_robotics}. 
Complementary work studies compact policies, efficient action tokenization, knowledge-preserving training, and asynchronous action-chunk execution~\citep{smolvla,fast,knowledge_insulation,rtc}.
Cross-embodiment policies must also reconcile heterogeneous control
interfaces. A physically interpretable unified action space is introduced
by~\citet{rdt1b}, whereas heterogeneous controls are mapped into semantically
aligned slots by~\citet{being_h05}.

This line of work primarily improves control representations, training scale,
and transfer across tasks and embodiments. Our low-level policy follows the
same generalist VLA paradigm. Our primary focus is complementary: making
high-level subtask generation an explicit, compute-scalable inference
problem.

\subsection{Hierarchical Robot Policies}

Hierarchical policies separate temporally extended decisions from continuous
control. Prior work grounds language-model decisions in learned robot
affordances~\citep{saycan}, represents action hierarchies through
language~\citep{rth}, and connects slower semantic reasoning to faster
visuomotor control through explicit commands, latent representations, or
staged inference~\citep{hirobot,helix,groot,gemini_robotics,pi05}.
Recent systems address long-horizon state tracking and adaptation more
explicitly. 
\citet{mem} combine short-term visual memory with long-term textual memory.
\citet{memer} retrieve task-relevant historical keyframes
before generating instructions for a low-level VLA, while~\citet{goal2skill}
combine structured memory, outcome verification, and reflection in a
high-level VLM and low-level VLA hierarchy. \citet{anticipation_vla}
adaptively generate recursive subgoals, whereas~\citet{vista} use a world
model as the high-level policy to generate textual and visual subgoal
sequences.

Among generalist hierarchical systems, one recent approach couples a semantic
high-level policy, a world model, and a low-level policy~\citep{pi07}. Its
world model generates subgoal images for a subtask
already produced by the high-level policy. In $\tau_0$-VLA, visual prediction
instead occurs before commitment and supports comparison among multiple
candidate subtasks through value-guided search and reflection. Thus, visual
prediction conditions how a generated subtask is executed in that framework,
whereas it informs which subtask to execute in $\tau_0$-VLA. Our high-level
policy also maintains a correctable execution memory and scales
decision-time computation through search width and depth.

\subsection{World Models and Test-Time Computation}

World models predict future states under candidate behavior and support
planning and policy learning~\citep{worldmodels,dreamerv3}. In robot
manipulation, generated images or videos have been used as goals for inverse
dynamics and low-level control~\citep{unipi,susie}, modeled jointly with
actions~\citep{gr2,worldvla}, and incorporated into search or iterative plan
revision~\citep{vlp}.
\citet{reflective_planning} iteratively revise a VLM plan using imagined
future states. Our search instead maintains multiple language-subtask
branches and prunes them with a dedicated value model before reflective
generation.
At the action level, additional inference is allocated through sampling,
verification, or value-guided selection~\citep{vgps,robomonkey,rover}.
\citet{forewarn} predict latent outcomes for low-level action plans and use a
VLM to select among them, while~\citet{vla_reasoner} perform
world-model-guided Monte Carlo tree search over action trajectories.
\citet{wla0} jointly predict textual subtasks, future images, and actions,
and their test-time scaling mode selects among action chunks using imagined
future frames and a value model.

Search has also been applied to high-level robot decisions. \citet{vine}
search structured scene-graph subgoal transitions, predict their success
probabilities from successful and failed demonstrations, and prune
low-feasibility branches before execution. \citet{seeing_farther} instead
search discrete manipulation plans with a learned critic, combining predicted
visual dynamics, value-guided beam search, confidence-based routing, and
multi-path reflection. Both are among the closest search-based mechanisms to
our work. By contrast, $\tau_0$-VLA searches open-ended language subtasks and
evaluates each candidate through its predicted terminal image. It conditions
subsequent decisions on a correctable execution memory, then uses a reflective
model to generate the final subtask from the retained branches.

\section{Preliminaries}
\label{sec:preliminaries}

We consider a robot task specified by a language instruction $\ell$. At
inference step $t$, a vision-language-action (VLA) policy maps the current
multi-view observation $o_t$, proprioceptive state $\mathbf{s}_t$, and language
command $c_t$ to an $H$-step action chunk
$\mathbf{a}_{t:t+H-1}$. It also receives textual control metadata $\eta$
specifying the embodiment, control mode, and whole-body configuration. Its
concrete text serialization is provided in
Appendix~\ref{app:unified-space}:
\begin{equation}
    \mathbf{a}_{t:t+H-1}
    =
    \pi_{\theta}(o_t,\mathbf{s}_t,c_t,\eta),
    \label{eq:vla_policy}
\end{equation}
where $\theta$ denotes the policy parameters and $H$ is the action-chunk
horizon. In direct execution, the full task instruction is used throughout the
episode, so $c_t=\ell$. The policy must therefore infer the current stage of
the task while generating the corresponding motor actions.

A hierarchical VLA separates high-level subtask generation from low-level
action generation. At each inference step, the high-level policy first
generates a subtask, and the low-level policy then generates actions
conditioned on it. Let $\mu$ denote the high-level policy, let
$\mathcal{M}_0$ denote the initial empty execution memory, and let
$z_0^\star=\varnothing$. At inference step $t$, $\mu$ forms the high-level
decision context
\begin{equation}
    h_t
    =
    \left(
    \ell,\,
    \mathcal{M}_{t-1},\,
    z_{t-1}^\star,\,
    o_t
    \right),
    \label{eq:high_level_context}
\end{equation}
where $\mathcal{M}_{t-1}$ is the carried execution memory and
$z_{t-1}^\star$ is the subtask generated at the preceding inference step. The
policy brings the execution memory up to date and generates the subtask for the
current observation:
\begin{equation}
    \left(
    \mathcal{M}_t,\,
    z_t^\star
    \right)
    =
    \mu(h_t).
    \label{eq:high_level_policy}
\end{equation}
Here $\mathcal{M}_t$ summarizes the execution history observed up to $o_t$ and
therefore does not yet contain the outcome of executing $z_t^\star$, while
$z_t^\star$ is the subtask generated for the current observation. The generated
subtask is used as the language command, $c_t=z_t^\star$, yielding
\begin{equation}
    \mathbf{a}_{t:t+H-1}
    =
    \pi_{\theta}(o_t,\mathbf{s}_t,z_t^\star,\eta).
    \label{eq:hierarchical_execution}
\end{equation}
After executing the action chunk, the resulting real observation is used at
the next inference step. At every inference step, the low-level policy
conditions on the generated subtask $z_t^\star$. If it remains the same as
$z_{t-1}^\star$, execution of the current subtask continues. If it changes,
the low-level policy begins executing the newly generated subtask.

The high-level policy can either commit to its direct proposal or invoke
additional test-time computation before producing $z_t^\star$. Thus,
the high-level policy determines \emph{which subtask is
currently appropriate}, while the low-level policy determines \emph{how to
execute it}.

\section{Method}
\label{sec:method}

\subsection{System Overview}

\begin{figure*}[t]
  \centering
  \includegraphics[width=0.98\textwidth]{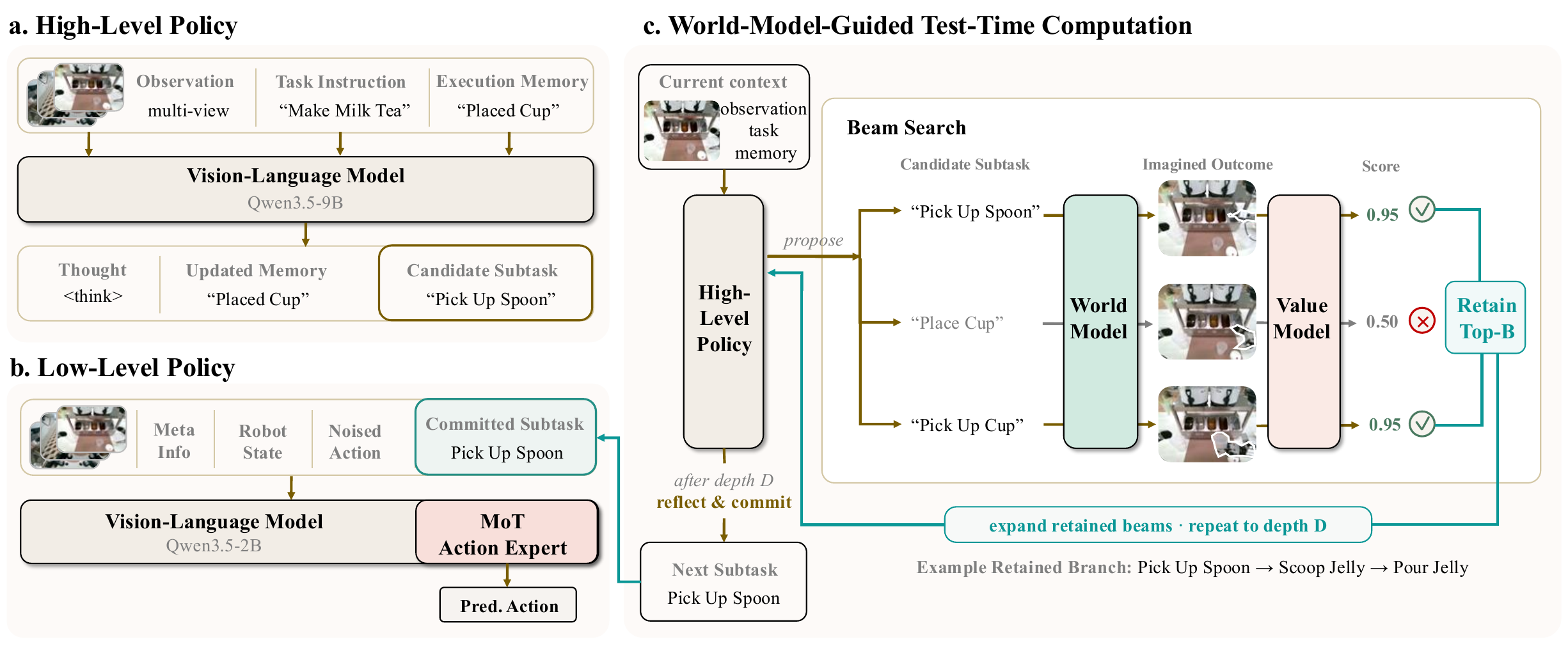}
  \caption{\textbf{The hierarchical $\tau_0$-VLA architecture.}
    (a) At a high-level inference step $t$, the proposal model $P$ conditions
    on the latest multi-view observation $o_t$, task instruction $\ell$,
    carried execution memory $\mathcal{M}_{t-1}$, and previously generated
    subtask $z_{t-1}^\star$. It produces observation-aligned memory
    $\mathcal{M}_t$ and a direct proposal $z_t^{\mathrm{dir}}$. (b) The
    low-level policy conditions on the generated subtask
    $z_t^\star$, multi-view observation $o_t$, proprioceptive state
    $\mathbf{s}_t$, and textual control metadata $\eta$. A
    vision-language backbone and Mixture-of-Transformers (MoT) action expert
    generate the action chunk $\mathbf{a}_{t:t+H-1}$ through conditional flow
    matching from a noisy action chunk. (c) On the TTC route, the proposal
    model is invoked $N$ times for each retained branch to generate $N$
    candidates. Given the branch's head-camera image and a candidate,
    the world model predicts the terminal head-camera image, and the value model
    assigns a candidate-quality score conditioned on the task instruction,
    candidate, and predicted image. Beam search globally retains the top-$B$
    branches by cumulative score and recursively expands them to depth $D$.
    The figure illustrates the root expansion with $N=3$ and $B=2$, where
    local and cumulative scores coincide. Deeper expansion is omitted for
    clarity. The reflective model then conditions on $\bar{h}_t$ and the final
    branch summaries $\mathcal{C}_t$ to generate the final subtask
    $z_t^\star$. This output may coincide with a retained proposal but is not
    restricted to the retained set and is passed to the low-level policy for
    execution.}
  \label{fig:method}
\end{figure*}

$\tau_0$-VLA consists of two components applied sequentially at each logical inference step. The \emph{high-level policy} maintains execution memory, decides when to invoke additional test-time computation, and generates the current subtask. The \emph{low-level policy} then maps the current observation and generated subtask to robot actions. This hierarchy supports long-horizon progress tracking and consequence-aware planning without modifying the low-level control interface. Figure~\ref{fig:method} provides an overview. The following subsections define both policies and then describe their joint inference procedure.

\subsection{High-Level Policy}

Unlike a conventional fixed-compute high-level policy which commits to a subtask through a single forward pass, our high-level policy $\mu$ generates the next subtask with a budget-adaptive test-time computation (TTC) procedure.
On uncertain predictions, TTC performs world-model-guided beam search over possible subtask sequences. It expands candidate branches, predicts and scores their visual outcomes, and passes the retained beam to a reflective model that generates the final subtask.
This procedure allows the policy to improve next-subtask prediction by allocating additional test-time computation.

Accordingly, the high-level policy $\mu$ consists of a proposal model $P$, a world model $\mathcal{W}$, a value model $V$, and a reflective model $F$.
Their inference interfaces are defined below.

\myparagraph{Proposal model.} At the beginning of inference step $t$, the
proposal model receives the context $h_t$ defined in
Section~\ref{sec:preliminaries}, updates the execution memory, and generates a
direct subtask proposal $z_t^{\mathrm{dir}}$:
\begin{equation}
    \left(z_t^{\mathrm{dir}},\mathcal{M}_t\right)=P(h_t).
    \label{eq:proposal-interface}
\end{equation}
The proposal model updates the memory from $\mathcal{M}_{t-1}$, the previously
generated subtask $z_{t-1}^\star$, and the current observation $o_t$. From
token confidences produced by the same forward pass, the adaptive router
computes $g_t\in\{0,1\}$. Here $g_t=0$
selects the fast route and $g_t=1$ invokes TTC. The routing rule is
defined in Appendix~\ref{app:adaptive-routing}.

\myparagraph{World model and value model.} For a generic candidate within beam
search, let $\tilde{o}$ denote a head-camera RGB image and $z$ the candidate
subtask. The world model predicts the terminal head-camera image, and the value
model scores that outcome:
\begin{equation}
    \hat{o}=\mathcal{W}(\tilde{o},z),
    \qquad
    v=V(\ell,z,\hat{o}).
    \label{eq:world-value-interfaces}
\end{equation}
Thus, the world model always operates on a single head-camera image. The value
model receives the global task instruction, candidate subtask, and predicted
terminal image, and returns a scalar candidate-quality score $v$.

\myparagraph{Test-time search.} When $g_t=1$, the high-level policy invokes the
\textsc{Search} operation in Algorithm~\ref{alg:system-inference}:
\begin{equation}
\mathcal{C}_t
=
\operatorname{Search}
\left(h_t,P,\mathcal{W},V,N,B,D\right).
\label{eq:search-interface}
\end{equation}
The operation performs beam search over candidate subtask sequences. The
positive integers $N$, $B$, and $D$ denote the branching factor, beam width,
and search depth. A branch $b$ stores a proposal context $h(b)$, an ordered
sequence of imagined subtasks $\rho(b)$, a cumulative score $S(b)$, and, for a
non-root branch, a terminal predicted image $\hat{o}(b)$. We initialize
$\mathcal{Q}_{t,0}$ with a single root branch
$b_{\mathrm{root}}$ whose context is $h_t$, whose path is the empty sequence
$\rho(b_{\mathrm{root}})=()$, and whose score is
$S(b_{\mathrm{root}})=0$.

At each depth $d\in\{1,\ldots,D\}$, we independently invoke the proposal model
$N$ times for every retained branch $b\in\mathcal{Q}_{t,d-1}$:
\begin{equation}
\left(z^{b,i}_{t,d},\mathcal{M}^{b,i}_{t,d}\right)
\sim P(\,\cdot\mid h(b)),
\qquad i=1,\ldots,N.
\end{equation}
Here $P(\,\cdot\mid h(b))$ denotes the proposal model's decoding distribution.
The sample index $i$ keeps repeated text proposals distinct and associates
each proposal with its branch-local memory. For the root branch,
$h(b_{\mathrm{root}})=h_t$, so the proposal model receives the current
multi-view observation $o_t$. For every non-root branch, the visual component
of $h(b)$ is $\hat{o}(b)$, the terminal head-camera image imagined for that
branch. Routing decisions from proposal calls inside search are ignored.

We write $b\oplus z$ for the child branch obtained by appending subtask $z$ to
branch $b$. Its path $\rho(b\oplus z)$ is the ordered sequence $\rho(b)$
followed by $z$. Let $\tilde{o}(b)$ be the head-camera image in $o_t$ when
$b=b_{\mathrm{root}}$ and $\hat{o}(b)$ otherwise. For each indexed proposal,
the world and value models compute
\begin{equation}
\begin{aligned}
\hat{o}\!\left(b\oplus z^{b,i}_{t,d}\right)
&=\mathcal{W}\!\left(\tilde{o}(b),z^{b,i}_{t,d}\right),\\
v\!\left(b\oplus z^{b,i}_{t,d}\right)
&=V\!\left(
\ell,z^{b,i}_{t,d},
\hat{o}\!\left(b\oplus z^{b,i}_{t,d}\right)
\right).
\end{aligned}
\end{equation}
The child retains the corresponding memory
$\mathcal{M}^{b,i}_{t,d}$ and predicted image. Its cumulative score and context
for the next expansion are
\begin{equation}
\begin{aligned}
S\!\left(b\oplus z^{b,i}_{t,d}\right)
&=S(b)+v\!\left(b\oplus z^{b,i}_{t,d}\right),\\
h\!\left(b\oplus z^{b,i}_{t,d}\right)
&=
\left(
\ell,\mathcal{M}^{b,i}_{t,d},z^{b,i}_{t,d},
\hat{o}\!\left(b\oplus z^{b,i}_{t,d}\right)
\right).
\end{aligned}
\end{equation}
All memories produced inside search are branch-local and never overwrite the
persistent execution memory $\mathcal{M}_t$.

At depth $d$, the indexed collection of all children is
\begin{equation}
\mathcal{A}_{t,d}
=
\left(
b\oplus z^{b,i}_{t,d}
\right)_{\substack{b\in\mathcal{Q}_{t,d-1}\\ i=1,\ldots,N}},
\qquad
\mathcal{Q}_{t,d}
=
\operatorname{Top}_{B}\!\left(\mathcal{A}_{t,d},S\right).
\end{equation}
Before pruning, $\mathcal{A}_{t,1}$ contains $N$ children, while each
subsequent expansion produces at most $BN$ children. The top-$B$ operation
ranks all children globally by cumulative score. Each retained child's
predicted image and branch-local memory define the context for its next
expansion. This process repeats until depth $D$.

The final beam is summarized by the indexed collection
\begin{equation}
\mathcal{C}_t=
\left(
\bigl(\rho(b),\hat{o}(b),S(b)\bigr)
\right)_{b\in\mathcal{Q}_{t,D}},
\end{equation}
where $\rho(b)$ is the imagined subtask path and $\hat{o}(b)$ is its terminal
predicted head-camera image.

\myparagraph{Reflective model.} At the end of the TTC route at inference step
$t$, the reflective model conditions on the retained branch summaries
$\mathcal{C}_t$ and the observation-aligned real context
$\bar{h}_t=(\ell,\mathcal{M}_t,z_{t-1}^\star,o_t)$. It generates the final
subtask passed to the low-level policy:
\begin{equation}
    z_t^\star=F(\bar{h}_t,\mathcal{C}_t).
    \label{eq:reflection-interface}
\end{equation}
Its output may reproduce a retained proposal but is not constrained to the
candidate set. The reflective model does not update persistent execution
memory.

\subsection{Low-Level Policy}
\label{sec:low-level-policy}

The low-level policy couples a vision-language backbone with a
Mixture-of-Transformers (MoT) action expert. At each full-attention layer, the
action tokens and backbone tokens interact through joint attention while being
processed by separately parameterized Transformer streams. The policy
conditions on multi-view observations, proprioceptive state, and a language
command. The action expert is trained with conditional flow matching to learn
a velocity field from noise to the distribution of action chunks. At
inference, this field is integrated to generate executable actions.

\myparagraph{Unified state and action space.}\label{sec:action-space}
We represent heterogeneous embodiments in a shared $40$-dimensional state and
action space. Per-sample state and action masks select the valid channels,
while the control metadata specifies the control parameterization. This
interface allows the same policy to support fixed-base and whole-body control
without embodiment-specific output heads.
Appendix~\ref{app:unified-space} provides the complete layout and action
encoding.

\myparagraph{Masked flow matching.}
Because different embodiments occupy different channels in the unified action
space, we mask both the flow path and its training objective. Let $d_a$ denote
the action dimension and let
$\mathbf{M}\in\{0,1\}^{d_a\times d_a}$ be a diagonal action-mask matrix, where
$[\mathbf{M}]_{ii}=1$ if channel $i$ is active and $[\mathbf{M}]_{ii}=0$
otherwise. The same matrix is applied to every action vector in the chunk.
For each $j\in\{0,\ldots,H-1\}$, let
$\mathbf{a}_{t+j},\boldsymbol{\epsilon}_{t+j}\in\mathbb{R}^{d_a}$, where
$\boldsymbol{\epsilon}_{t+j}\sim
\mathcal{N}(\mathbf{0},\mathbf{I}_{d_a})$ and $\mathbf{I}_{d_a}$ is the
$d_a$-dimensional identity matrix. Following the linear Gaussian flow path
of~\citet{pi0}, we use its algebraically equivalent reverse-time
reparameterization to match the task-specific checkpoints evaluated in this
work. A flow time
$\tau\in[0,1]$ is shared across the chunk, with $\tau=1$ denoting noise and
$\tau=0$ denoting a clean action. The interpolated action and target velocity
are
\begin{equation}
    \begin{aligned}
    \mathbf{a}_{t+j}^{\tau}
    &=
    \tau\mathbf{M}\boldsymbol{\epsilon}_{t+j}
    +(1-\tau)\mathbf{M}\mathbf{a}_{t+j},\\
    \mathbf{u}_{t+j}
    &=
    \mathbf{M}
    (\boldsymbol{\epsilon}_{t+j}-\mathbf{a}_{t+j}).
    \end{aligned}
    \label{eq:masked-flow-path}
\end{equation}
Given the complete noisy chunk, the action expert jointly predicts the
velocities of all $H$ action tokens. We supervise only active channels and
project inactive channels before each velocity-field evaluation and at the
final output. Appendix~\ref{app:unified-space} provides the full loss and
sampling details.

\subsection{System Inference}
\label{sec:system-inference}

At inference, the high-level policy and low-level policy form a closed loop.
At each logical inference step, the proposal model updates the execution
memory and generates $z_t^{\mathrm{dir}}$. When $g_t=0$, this direct proposal
becomes the final subtask $z_t^\star$. When $g_t=1$, test-time search produces
$\mathcal{C}_t$, and the reflective model generates $z_t^\star$ from the
observation-aligned context and retained branch summaries. The low-level policy
then generates and executes an action chunk conditioned on $z_t^\star$. The
resulting real observation is incorporated into memory at the next inference
step.

Algorithm~\ref{alg:system-inference} presents these logical dependencies as a
sequential procedure. In deployment, the high-level and low-level policies are
pipelined asynchronously, as detailed in Appendix~\ref{app:async-serving}. The
same routing statistics are used across tasks, with task-specific thresholds
calibrated on held-out data. In direct-execution mode, the system bypasses the
high-level policy and conditions the low-level policy directly on $\ell$.

\begin{algorithm}[t]
\caption{Closed-loop system inference}
\label{alg:system-inference}
\begin{algorithmic}[1]
\Require Task instruction $\ell$, models $P,\mathcal{W},V,F,\pi_\theta$
\Require Control metadata $\eta$, action horizon $H$, branching factor $N$, beam width $B$, depth $D$
\State Initialize $\mathcal{M}_0\gets\varnothing$ and
  $z_0^\star\gets\varnothing$
\For{$t=1,2,\ldots$ until task termination}
  \State Observe $o_t$ and $\mathbf{s}_t$
  \State $h_t\gets(\ell,\mathcal{M}_{t-1},z_{t-1}^\star,o_t)$
  \State $(z_t^{\mathrm{dir}},\mathcal{M}_t)\gets P(h_t)$
  \State Compute $g_t$ from the proposal token confidences
  \State $\bar{h}_t\gets(\ell,\mathcal{M}_t,z_{t-1}^\star,o_t)$
  \If{$g_t=0$}
    \State $z_t^\star\gets z_t^{\mathrm{dir}}$ \Comment{fast route}
  \Else
    \State $\mathcal{C}_t\gets
      \Call{Search}{h_t,P,\mathcal{W},V,N,B,D}$
    \State $z_t^\star\gets F(\bar{h}_t,\mathcal{C}_t)$
  \EndIf
  \State $\mathbf{a}_{t:t+H-1}\gets
    \pi_\theta(o_t,\mathbf{s}_t,z_t^\star,\eta)$
  \State Execute action chunk $\mathbf{a}_{t:t+H-1}$
\EndFor
\end{algorithmic}
\end{algorithm}









\section{Training}
\label{sec:training}

\subsection{Data Sources}
\label{sec:data-sources}

\myparagraph{Low-level policy data.}The low-level policy is trained on $40{,}115$ hours of heterogeneous real-world
data spanning fixed-base, mobile, and bimanual embodiments. The corpus
combines human demonstrations, autonomous policy rollouts, and UMI-style
recordings, covering manipulation, navigation, and whole-body coordination
across diverse robot morphologies and control interfaces. We interleave these
trajectories with multimodal vision-language data for instruction following,
visual grounding, spatial and depth reasoning, and robot-centric perception.
This co-training mixture preserves the semantic and visual capabilities of the
VLM backbone during action learning. Appendix~\ref{app:low-level-data} provides
further details on the data composition and processing.

\myparagraph{High-level policy data.}Supervision for the high-level policy is derived automatically from existing
task, stage, and executable-subtask annotations together with segmented
multi-view demonstrations. These sources supervise reasoning over task
progress, execution-memory updates, and subtask generation. We additionally
construct memory-perturbed examples that teach the policy to recover when its
execution history lags behind, runs ahead of, or otherwise conflicts with the
observed state. Subtask-aligned start--end frames supervise the world model.
This construction requires no additional per-sample human labeling.
Appendix~\ref{app:hl-data} describes the annotation, filtering, and quality
control pipeline.






\subsection{Low-Level Policy Training}

We initialize the low-level vision-language backbone from
Qwen3.5-2B~\citep{qwen35} and attach the MoT action expert described in
Section~\ref{sec:low-level-policy}. The resulting policy maps the current
observation and language command to robot actions in the unified action space,
as defined in Eq.~\eqref{eq:vla_policy}. We train it in three successive
stages.

\myparagraph{Stage 1: Knowledge-isolated co-training.}
Following~\citet{knowledge_insulation}, we jointly train on multimodal and
robot-action data with knowledge isolation (KI). The MoT action expert attends
to the vision-language backbone, while action-loss gradients are blocked at
the backbone interface. This allows the action expert to learn
control-relevant representations without prematurely perturbing the
pretrained backbone, with multimodal supervision stabilizing training.

\myparagraph{Stage 2: End-to-end co-training.}We remove KI and optimize the full model end to end, retaining multimodal data as auxiliary supervision. This allows action gradients to adapt the backbone toward representations better suited for robot control and strengthens the coupling between perception, language, and action prediction.

\myparagraph{Stage 3: Task-specific adaptation.}For each deployment task, we fine-tune the policy on a small set of task-specific demonstrations, adapting it to the target embodiment, camera viewpoints, object configurations, and success criteria.

\subsection{High-Level Policy Training}

The proposal, value, and reflective models are independently fine-tuned from
the same robot-pretrained VLM checkpoint initialized from
Qwen3.5-9B~\citep{qwen35}. The world model is initialized from
Step1X-Edit~\citep{step1x_edit} and trained separately. We describe the specific supervision for each model below.

\myparagraph{Proposal model.}The proposal model generates candidate subtasks while maintaining and updating
execution memory. We train it on both aligned execution histories and automatically
perturbed histories that lag behind, run ahead of, or misrepresent progress
after a failure. The corrected targets teach the model to reconcile memory with
visual evidence before generating the next subtask. The perturbation types,
recovery targets, and sampling mixture are detailed in
Appendix~\ref{app:hl-data}.

\myparagraph{World model.}The world model is trained to predict the visual
outcome of a candidate subtask at completion. Each training example pairs the
head-camera RGB observations at the beginning and end of an annotated subtask
segment, conditioned on the corresponding subtask instruction. These
subtask-aligned transitions are drawn from real trajectories.

Starting from real observations, we construct multi-step simulated rollouts by recursively alternating subtask proposal with $P$ and visual outcome prediction with $\mathcal{W}$, using each predicted observation to condition the next proposal step.
These simulated rollouts form an offline dataset used to train the value and reflective models described below.

\myparagraph{Value model.}We formulate value prediction as a multiple-choice
VQA task. Conditioned on the global instruction, candidate subtask, and its
imagined terminal image, the model predicts one of five ordinal quality levels,
from \emph{clearly wrong} to \emph{clearly correct}, mapped to scalar values in
$[0.05,0.95]$.
Training pairs each sampled state with a candidate subtask from offline
rollouts generated by repeatedly applying the proposal and world models. The
candidates are graded against the ground-truth next step, aligning training
with the proposals that the value model scores during search.

\myparagraph{Reflective model.}We train the reflective model with the same
observation-aligned context and retained-branch summaries used at inference.
The candidate plans come from the same offline rollouts together with their
predicted future states and scores. Each set is paired with the ground-truth
next subtask as an autoregressive target. This supervision teaches the model
to ground its generation in visual evidence and repair flawed candidate plans
rather than simply copy the first proposal.

\section{Experiments}

Our evaluation separates high-level decision quality, long-horizon system
performance, and short-horizon execution across embodiments. At the high level,
we evaluate next-subtask prediction, test-time computation, and the
contribution of execution memory. We then evaluate the hierarchical system on
four long-horizon tasks, study test-time computation on Book Organization task, and
evaluate the VLA through direct execution on the shorter Collect Laundry and
Tidy Makeup Table tasks.

\begin{figure*}[t]
  \centering
  \includegraphics[width=1.0\textwidth]{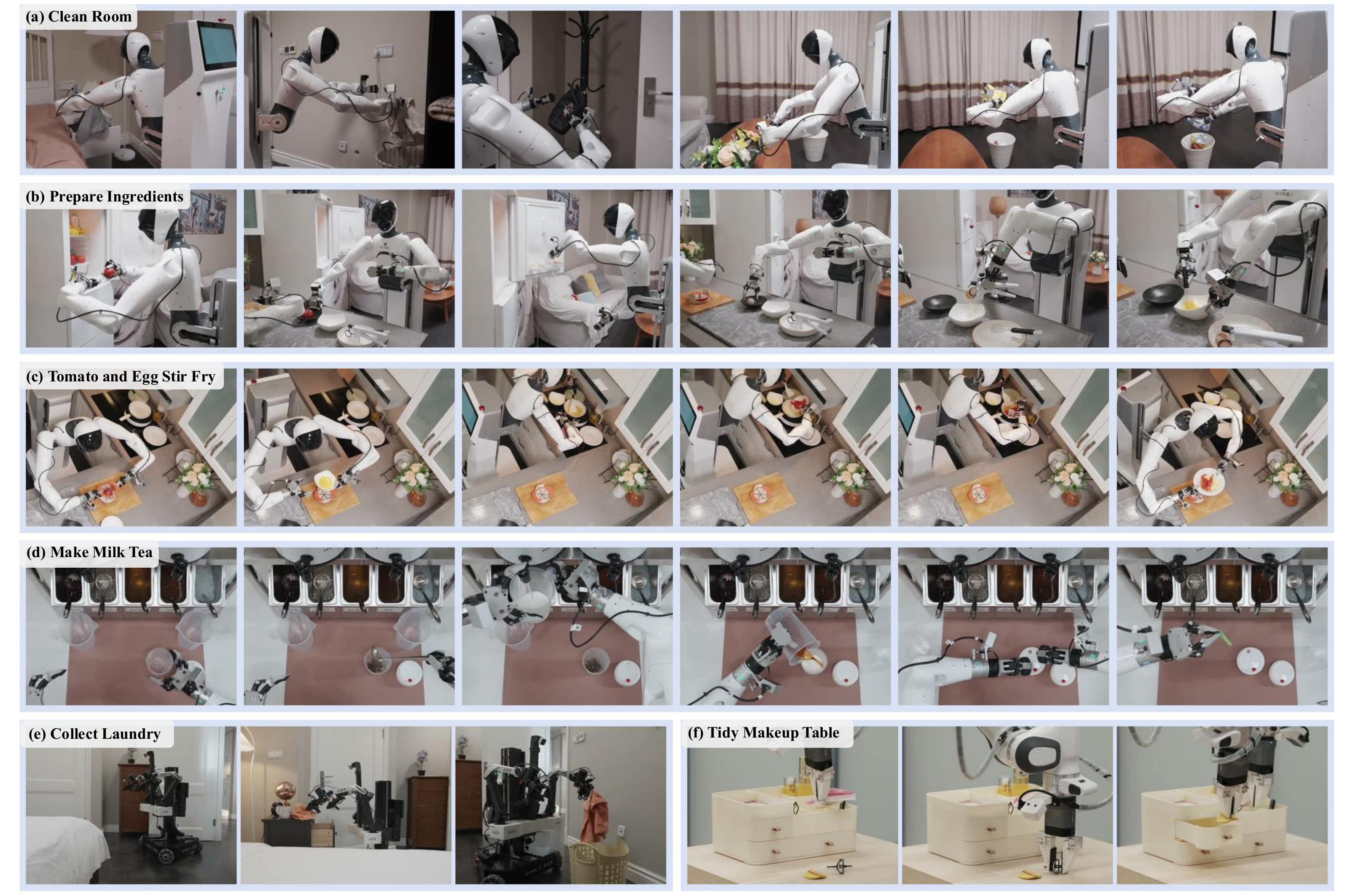}
  \caption{\textbf{Representative physical-robot evaluation tasks.}
  (a) Clean Room requires collecting two dirty garments, placing them in a
  laundry basket, hanging a handbag, handing a blanket to a person, and
  disposing of table trash. (b) Prepare Ingredients requires retrieving a
  tomato and an egg, then cracking and stirring the egg while returning the
  tools. (c) Tomato and Egg Stir Fry requires chopping and transferring the
  tomatoes, cooking and seasoning the ingredients, plating the dish, and
  returning the cookware and utensils.
  (d) Make Milk Tea requires adding toppings, pouring milk and tea, sealing the
  cup, and inserting a straw. (e) Collect Laundry requires transferring a
  T-shirt from the bedside table to a laundry basket. (f) Tidy Makeup Table
  comprises three separately scored instruction-following groups that require
  different object selections and action sequences from matched visual states.}
  \label{fig:demo_illus}
\end{figure*}

\subsection{Evaluation Setup}

\myparagraph{Robot platforms.}
We evaluate on AGIBOT G1 for the four primary long-horizon tasks, ARX AC One
for Book Organization and Collect Laundry, and a bimanual Franka Research 3
setup for Tidy Makeup Table.
All platforms provide multi-view RGB observations and proprioceptive state and
map their controls into the unified action space. Hardware and sensing
details are deferred to Appendix~\ref{app:robot-platforms}.

\myparagraph{Task suite.}
The evaluation covers six household tasks and one benchmark comprising three
separately evaluated instruction-following task groups. Clean Room, Prepare
Ingredients, Tomato and Egg Stir Fry, and Collect Laundry require mobile
manipulation. Make Milk Tea, Book Organization, and the three Tidy Makeup Table
groups require manipulation without base motion. We use Tidy Makeup Table for
instruction-following evaluation. Prepare Ingredients and Tomato and Egg Stir
Fry form a collaborative cooking workflow for the same tomato and egg dish. One
robot prepares the ingredients, and another completes the cooking stage. We
therefore evaluate and report the two stages as separate tasks.
\begin{itemize}
  \item \textbf{Clean Room (25 steps).} The robot enters the bedroom, retrieves two dirty garments from the nightstand and bed, and places them in a laundry basket. It then hangs a handbag on a clothes rack, retrieves a blanket and hands it to a person, leaves the bedroom, and disposes of snack-bag trash from the coffee table. A typical successful rollout lasts approximately $8$\,min.
  \item \textbf{Prepare Ingredients (14 steps).} The robot moves between the refrigerator and preparation table, opens the refrigerator, retrieves a tomato and an egg, and closes the refrigerator. It then places the ingredients in bowls, cracks the egg with an egg cracker, stirs the egg mixture, and returns both tools to a porcelain plate. A typical successful rollout lasts approximately $4$\,min.
  \item \textbf{Tomato and Egg Stir Fry (22 steps).} The robot chops the tomatoes and transfers the tomatoes and prepared egg mixture to the induction cooker. It turns on the cooker, adds oil and the egg mixture, stir-fries the eggs, adds the tomatoes and salt, finishes stir-frying, plates the dish, returns the cookware and utensils, turns off the cooker, places the finished dish on the preparation table, and returns both arms to the home position. A typical successful rollout lasts approximately $10$\,min.
  \item \textbf{Make Milk Tea (13 steps).} The robot places a cup at the preparation station, adds two toppings, pours milk and tea in sequence, seals the cup with a lid, and inserts a straw. A typical successful rollout lasts approximately $3$\,min.
  \item \textbf{Collect Laundry (5 steps).} The robot searches for and moves to the drawer cabinet, picks up the dirty T-shirt from the bedside table with its left arm, closes the cabinet door with its right arm, searches for and moves to the laundry basket, and places the T-shirt in the basket with its left arm. A typical successful rollout lasts approximately $1$\,min.
  \item \textbf{Tidy Makeup Table (three task groups).} This benchmark pairs matched visual states with instructions that specify different target objects, active arms, action sequences, and destinations. We score three groups independently. In \emph{Cotton Pad} (2 steps), the left arm picks up the cotton pad and places it in its designated compartment. In \emph{Eyelash Curler} (2 steps), the left arm picks up the eyelash curler and places it in its designated compartment. In \emph{Makeup Puff} (4 steps), both arms open the drawer, the left arm picks up the makeup puff from the tabletop and places it inside, and both arms close the drawer. Executing all eight steps across the three task groups takes approximately $30$\,s in total.
  \item \textbf{Book Organization (3 steps).} The robot uses its two arms to reorder four books of different heights across four fixed slots, swapping any pair of books at each step until they are arranged from tallest to shortest. A typical successful rollout lasts approximately $1$\,min.
  We evaluate two initial-state regimes while keeping the task objective and action space fixed: in-domain initial arrangements appear in the training data, whereas out-of-domain (OOD) initial arrangements are not observed during training. 
  The OOD setting therefore evaluates the high-level policy's ability to predict appropriate swaps from observations induced by unseen initial configurations.
  
\end{itemize}

\myparagraph{Evaluation protocols.}
We use complementary protocols for high-level decision making and
physical-robot execution.

\par\noindent\textit{High-level policy evaluation.}\quad
We isolate high-level decision making on annotated subtask-boundary samples and
report next-subtask prediction accuracy and inference time. All methods within
each comparison receive the same inputs and output format. Evaluation-set
construction, variant-specific inputs, and judging details are provided in
Appendix~\ref{app:high-level-eval-protocol}.

\par\noindent\textit{Physical-robot evaluation.}\quad
We report \emph{success rate} (SR) and \emph{progress} using the same
task-specific terminal conditions for all methods. SR measures the fraction
of successful trials. A trial is successful only when every required milestone
is completed. Skipping a required milestone makes the trial unsuccessful.
Progress measures how much of the task is completed using annotated
milestones. Each milestone receives full, half, or zero credit based on
completion quality, and dependent milestones count only after their required
earlier steps are completed. Exact scoring rules, trial durations, reset
procedures, and termination conditions are provided in
Appendix~\ref{app:physical-eval-protocol}.

\subsection{Overall System Evaluation}

Table~\ref{tab:main} reports the four long-horizon tasks: Clean Room,
Prepare Ingredients, Tomato and Egg Stir Fry, and Make Milk Tea. These tasks require
navigation, object search, manipulation, state tracking, and recovery across
multiple stages. For the controlled comparison of our two variants, the
observation and action interfaces and the low-level policy are fixed.
Standalone $\tau_0$-VLA conditions the low-level policy on the full task
instruction. The hierarchical system instead supplies bounded subtasks from
the high-level policy. Both variants in this comparison run without beam
search, which is evaluated separately in the test-time computation experiments.
\begin{table*}[t]
\centering
\caption{\textbf{Long-horizon task performance.} Each
method--task setting uses 10 independently collected physical-robot trials. SR
reports successful trials as $x/10$, and Progress is the normalized
milestone-completion score. Avg. is the unweighted mean of the four task-level rates.
The first four rows use direct execution. The final row uses the Hierarchical
System with Plan Once and no beam search.}
\label{tab:main}
\footnotesize
\setlength{\tabcolsep}{1.2pt}
\renewcommand{\arraystretch}{1.15}
\begin{tabular*}{0.98\textwidth}{@{\extracolsep{\fill}}c*{10}{c}@{}}
\toprule
Method
& \multicolumn{2}{c}{Clean Room}
& \multicolumn{2}{c}{Prepare Ingredients}
& \multicolumn{2}{c}{Tomato and Egg Stir Fry}
& \multicolumn{2}{c}{Make Milk Tea}
& \multicolumn{2}{c}{Avg.} \\
\cmidrule(lr){2-3}\cmidrule(lr){4-5}\cmidrule(lr){6-7}\cmidrule(lr){8-9}\cmidrule(l){10-11}
& \multicolumn{1}{c}{SR $\uparrow$} & \multicolumn{1}{c}{Progress $\uparrow$}
& \multicolumn{1}{c}{SR $\uparrow$} & \multicolumn{1}{c}{Progress $\uparrow$}
& \multicolumn{1}{c}{SR $\uparrow$} & \multicolumn{1}{c}{Progress $\uparrow$}
& \multicolumn{1}{c}{SR $\uparrow$} & \multicolumn{1}{c}{Progress $\uparrow$}
& \multicolumn{1}{c}{SR $\uparrow$} & \multicolumn{1}{c}{Progress $\uparrow$} \\
\midrule
$\mathrm{GR00T}$~N1.7~\citep{groot}
& 0/10 & 59.80\% & 1/10 & 68.57\% & 0/10 & 24.32\%
& 0/10 & 28.46\% & 2.50\% & 45.29\% \\
LingBot-VLA~\citep{lingbot}
& 0/10 & 66.60\% & 0/10 & 35.00\% & 0/10 & 12.27\%
& 0/10 & 63.85\% & 0.00\% & 44.43\% \\
$\pi_{0.5}$~\citep{pi05}
& 4/10 & 86.20\% & 2/10 & 73.93\% & 0/10 & 49.77\%
& 3/10 & 82.31\% & 22.50\% & 73.05\% \\
$\tau_0$-VLA
& 4/10 & 92.80\% & 2/10 & 66.43\% & 0/10 & 65.00\%
& \textbf{5/10} & \textbf{96.15\%} & 27.50\% & 80.10\% \\
$\tau_0$-VLA (Hierarchical System, Plan Once)
& \textbf{5/10} & \textbf{94.80\%}
& \textbf{4/10} & \textbf{82.86\%}
& \textbf{4/10} & \textbf{81.82\%}
& \textbf{5/10} & 91.92\%
& \textbf{45.00\%} & \textbf{87.85\%} \\
\bottomrule
\end{tabular*}
\end{table*}

\myparagraph{Clean Room.}
Performance on Clean Room benefits clearly from explicit execution memory. The
hierarchical system retains progress across room transitions, whereas failures
without this memory concentrate on handbag hanging and the later room-tidying
stages.

\myparagraph{Prepare Ingredients.}
Most failures in Prepare Ingredients trials arise during egg pickup, cracking,
or stirring. Because these actions are prerequisites for later preparation
stages, an early error blocks substantial downstream progress. Tracking
completed stages is therefore particularly valuable on this task.

\myparagraph{Tomato and Egg Stir Fry.}
The decisive bottleneck in Tomato and Egg Stir Fry is seasoning. Adding salt
causes little visible change, so the current observation alone cannot reliably
reveal whether this step has already been completed. Direct-execution policies
therefore tend to add salt repeatedly or skip it entirely, either of which
violates the success criterion. The hierarchical system resolves this ambiguity
by recording seasoning progress explicitly.

\myparagraph{Make Milk Tea.}
Make Milk Tea presents a complementary regime in which both $\tau_0$-VLA
variants already complete most of the sequence reliably. Both achieve an SR of
5/10 and more than 91\% progress, outperforming the external baselines.
Their remaining failures occur during lid attachment or straw insertion, after
the preceding preparation stages have been completed. This pattern identifies
final contact-rich manipulation as the main remaining bottleneck. When TTC is
enabled, the hierarchical policy further improves to an SR of 7/10 and 95.38\%
progress, as reported in Table~\ref{tab:test_time_compute_success}.

\subsection{Evaluation Across Embodiments}

We evaluate the low-level policy on two additional embodiments using Collect
Laundry and Tidy Makeup Table. Because these
tasks contain only two to five steps, each method executes the full task
instruction without task decomposition, execution memory, or test-time search.
This setting evaluates low-level control and language-conditioned manipulation
across embodiments, separately from the high-level policy used in the
long-horizon tasks.
Table~\ref{tab:additional_results} reports the results.
\begin{table*}[t]
\centering
\caption{\textbf{Direct-execution performance across embodiments.} Tidy Makeup
Table comprises three independently scored instruction-following groups. All
methods execute the full task instruction without a high-level policy. Each
task or task group is evaluated over 10 trials. SR reports successful trials
as $x/10$, and Progress is the normalized milestone-completion score.}
\label{tab:additional_results}
\footnotesize
\setlength{\tabcolsep}{2pt}
\renewcommand{\arraystretch}{1.15}
\begin{tabular}{@{}>{\centering\arraybackslash}p{0.20\linewidth}*{8}{>{\centering\arraybackslash}p{0.07\linewidth}}@{}}
\toprule
\multicolumn{1}{c}{Method}
& \multicolumn{2}{c}{Collect Laundry}
& \multicolumn{6}{c}{Tidy Makeup Table} \\
\cmidrule(lr){2-3}\cmidrule(l){4-9}
& \multicolumn{2}{c}{T-shirt}
& \multicolumn{2}{c}{Cotton Pad}
& \multicolumn{2}{c}{Eyelash Curler}
& \multicolumn{2}{c}{Makeup Puff} \\
\cmidrule(lr){2-3}\cmidrule(lr){4-5}\cmidrule(lr){6-7}\cmidrule(l){8-9}
& \multicolumn{1}{c}{SR $\uparrow$} & \multicolumn{1}{c}{Progress $\uparrow$}
& \multicolumn{1}{c}{SR $\uparrow$} & \multicolumn{1}{c}{Progress $\uparrow$}
& \multicolumn{1}{c}{SR $\uparrow$} & \multicolumn{1}{c}{Progress $\uparrow$}
& \multicolumn{1}{c}{SR $\uparrow$} & \multicolumn{1}{c}{Progress $\uparrow$} \\
\midrule
$\mathrm{GR00T}$~N1.7~\citep{groot}
& 4/10 & 76.00\% & \textbf{10/10} & 87.50\% & 8/10 & 77.50\% & 7/10 & 52.50\% \\
LingBot-VLA~\citep{lingbot}
& 2/10 & 35.00\% & 9/10 & 67.50\% & 3/10 & 22.50\% & 3/10 & 33.75\% \\
$\pi_{0.5}$~\citep{pi05}
& 9/10 & 88.00\% & 9/10 & 85.00\% & 8/10 & 85.00\% & 7/10 & 73.75\% \\
$\tau_0$-VLA
& \textbf{10/10} & \textbf{97.00\%}
& \textbf{10/10} & \textbf{95.00\%}
& \textbf{9/10} & \textbf{92.50\%}
& \textbf{10/10} & \textbf{95.00\%} \\
\bottomrule
\end{tabular}
\end{table*}

\myparagraph{Collect Laundry.}
Collect Laundry evaluates mobile manipulation by the low-level policy. The
observed failures concentrate on T-shirt grasping and navigation
around the bed: LingBot-VLA sometimes fails to lift the T-shirt,
while GR00T collides with the foot of the bed and terminates the rollout.

\myparagraph{Tidy Makeup Table.}
The Tidy Makeup Table benchmark evaluates instruction following under matched
visual states. The language instruction changes the target object,
active arm, action order, and destination, so the observation alone does not
determine the correct action sequence. Cotton Pad and Eyelash Curler emphasize
instruction-conditioned object selection and placement, whereas Makeup Puff
adds bimanual drawer manipulation. GR00T occasionally pauses during drawer
motion or selects the makeup puff instead of the eyelash curler. LingBot-VLA
generally follows the specified arm but tends to select visually nearby objects
rather than the instructed target. $\pi_{0.5}$ sometimes releases and re-grasps
the makeup puff, closes the drawer less smoothly, or omits the cotton pad.
These failures distinguish language-grounding errors from low-level execution
inefficiencies: selecting or omitting the wrong object changes task completion,
whereas re-grasping and hesitant drawer motion reduce execution efficiency even
when the intended final state is reached.

\subsection{Test-Time Computation Experiments}

We conduct several experiments to quantify the gains in subtask-prediction accuracy achieved by test-time computation (TTC) and assess whether these gains lead to higher success rates in closed-loop real-robot evaluation.
Our TTC experiments cover the following tasks: Make Milk Tea, Book Organization, and Clean Room.
We evaluate Book Organization both in-domain and out-of-domain (OOD) initial book arrangements: the former appears in the training data, whereas the latter does not.
The OOD setting evaluates the high-level policy's performance when planning from unseen initial observations.
We first measure next-subtask prediction accuracy in an open-loop setting, then evaluate TTC in closed-loop real-robot execution, and finally examine the relationship between computational cost and prediction accuracy.

\begin{figure}[htbp]
  \centering
  \includegraphics[width=\columnwidth]{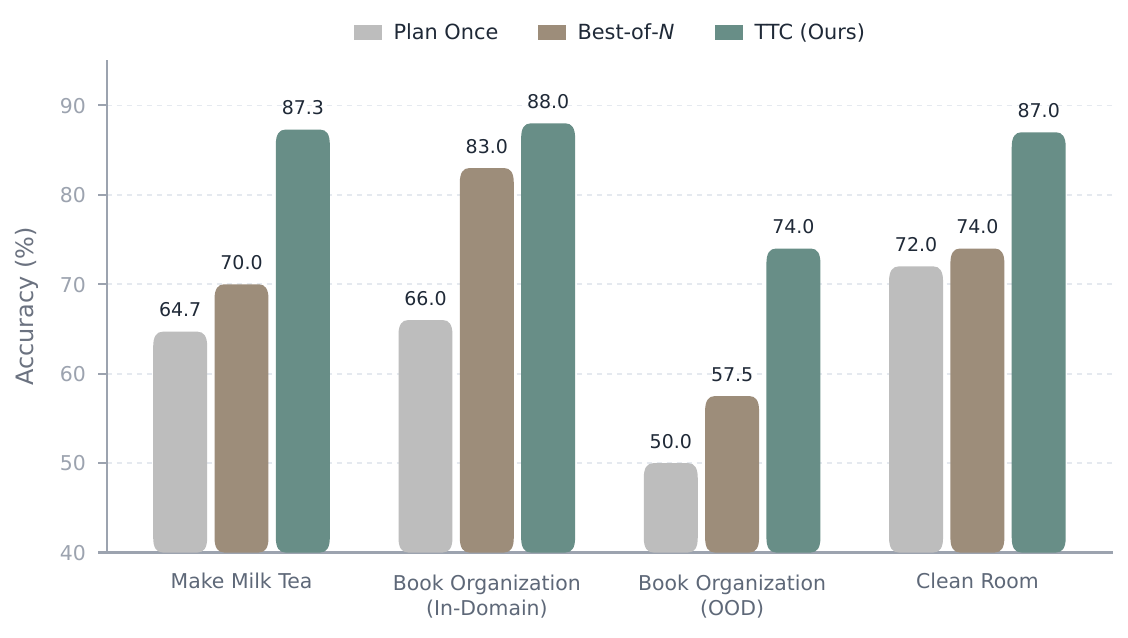}
  \caption{\textbf{Next-subtask prediction accuracy under different high-level inference methods.}
    We compare Plan Once, Best-of-$N$, and TTC (Ours) across four evaluation settings:
    Make Milk Tea, Book Organization (In-Domain), Book Organization (OOD), and
    Clean Room.}
  \label{fig:llm_judge_accuracy}
\end{figure}

\begin{table}[htbp]
\centering
\caption{\textbf{Closed-loop physical-robot performance with test-time computation.}
Each entry uses 10 independently collected trials. Book Organization uses
shuffled initial arrangements and is reported without the in-domain and OOD
split used in the open-loop evaluation. SR denotes task success rate, and
Progress is the normalized milestone-completion score.}
\label{tab:test_time_compute_success}
\footnotesize
\setlength{\tabcolsep}{1.5pt}
\renewcommand{\arraystretch}{1.15}
\begin{tabular*}{\columnwidth}{@{\extracolsep{\fill}}c*{6}{c}@{}}
\toprule
\multirow{2}{*}{Method}
& \multicolumn{2}{c}{Make Milk Tea}
& \multicolumn{2}{c}{Book Organization}
& \multicolumn{2}{c}{Clean Room} \\
\cmidrule(lr){2-3}\cmidrule(lr){4-5}\cmidrule(l){6-7}
& \multicolumn{1}{c}{SR $\uparrow$} & \multicolumn{1}{c}{Progress $\uparrow$}
& \multicolumn{1}{c}{SR $\uparrow$} & \multicolumn{1}{c}{Progress $\uparrow$}
& \multicolumn{1}{c}{SR $\uparrow$} & \multicolumn{1}{c}{Progress $\uparrow$} \\
\midrule
Plan Once & 5/10 & 91.92\% & 6/10 & 66.67\% & 5/10 & 94.80\% \\
TTC & \textbf{7/10} & \textbf{95.38\%} & \textbf{9/10} & \textbf{93.33\%} & \textbf{7/10} & \textbf{97.60\%} \\
\bottomrule
\end{tabular*}
\end{table}

\begin{figure*}[t]
  \centering
  \includegraphics[width=0.80\textwidth]{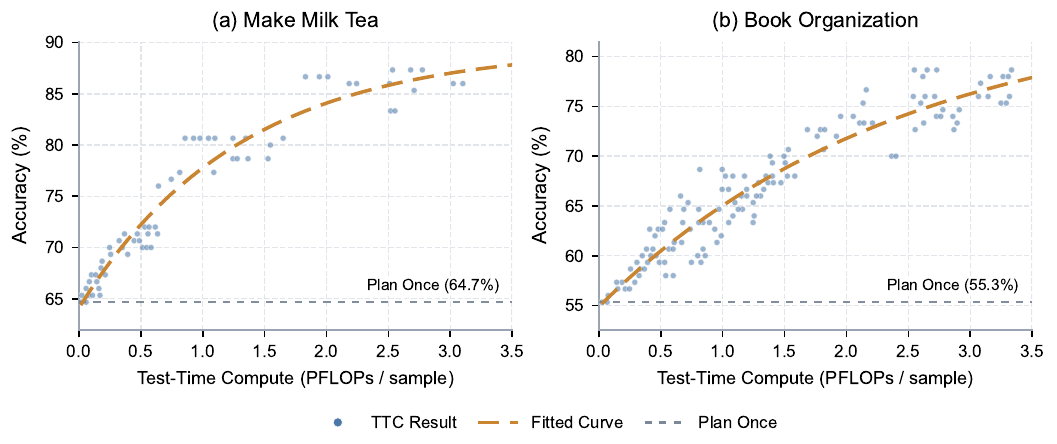}
  \caption{\textbf{Relationship between computational cost and subtask-prediction accuracy.}
    The accuracy increases with additional computation for the Make Milk Tea and Book Organization tasks, shown in panels (a) and (b), respectively.
    Each point represents an experimental result obtained at a different computational cost. The orange dashed curves show saturation fits to these results, and the gray dashed lines indicate the Plan Once baselines.}
  \label{fig:test_time_scaling}
\end{figure*}

\myparagraph{Open-loop Subtask Prediction.}
Following the high-level policy evaluation protocol above, we evaluate next-subtask
prediction at different stages of each task. We compare TTC against two
baselines. \textbf{Plan Once} makes one high-level prediction at each decision
point without test-time search. It remains a variant of the high-level policy
and is distinct from direct execution, which bypasses the high-level policy.
\textbf{Best-of-$N$} samples $N$ candidates, predicts the next
observation for each candidate using the same world model as TTC, and selects
the candidate assigned the highest score by the same value model.

As shown in Figure~\ref{fig:llm_judge_accuracy}, TTC achieves the highest accuracy across all four evaluation settings, demonstrating its effectiveness in improving next-subtask prediction.
Take the OOD Book Organization setting as an example, where TTC achieves $74.0\%$ accuracy, compared with $50.0\%$ for Plan Once and $57.5\%$ for Best-of-$N$.
In the OOD setting, the initial book arrangements are absent from the training data, placing the corresponding observations outside the high-level policy's training distribution.
Under this distribution shift, directly predicting the next subtask with the fine-tuned VLM is more prone to error.
Instead, TTC recursively expands candidate branches by using the world model to
predict future observations and the value model to score the resulting imagined
branches. The reflective model then uses the retained branches as context to
generate the current subtask.
Unlike direct VLM prediction based solely on patterns learned during fine-tuning, TTC evaluates candidate consequences at decision time, resulting in improved next-subtask prediction accuracy.
Although Best-of-$N$ evaluates sampled candidates using the same world and value models, it performs only one-step selection without TTC's multi-step expansion and reflective commitment.
Consequently, its accuracy gains relative to Plan Once are consistently smaller than those achieved by TTC across the evaluated settings.

\myparagraph{Closed-loop Real-robot Evaluation.}
We further assess whether TTC leads to higher task success rates in closed-loop real-robot evaluation.
We deploy the high-level policy with and without TTC while keeping the low-level policy fixed.
As reported in Table~\ref{tab:test_time_compute_success}, TTC improves both task progress and success rate across all evaluated tasks.
This benefit is particularly important for tasks such as Book Organization, which do not admit a fixed execution plan.
These results show that the gains in next-subtask prediction accuracy achieved by TTC lead to more effective and efficient closed-loop task execution.

\myparagraph{Computational Cost vs. Subtask-Prediction Accuracy.}
Finally, we conduct an analysis of the relationship between test-time computational cost and subtask-prediction accuracy.
Figure~\ref{fig:test_time_scaling} reports this relationship for the Make Milk Tea and Book Organization tasks.
Each point represents an experimental result with a different computational cost, and the orange dashed curves provide approximate saturation fits to the observations.
Specifically, we fit a saturating exponential function.
The parameters are estimated by least squares using all experimental observations.
Accuracy rises rapidly when computation is increased from a small budget, showing that additional search and reflection substantially improve subtask prediction in the low-compute regime.
The marginal gain then gradually decreases and the fitted curves approach a plateau.
This trend indicates that TTC provides a favorable compute--accuracy trade-off at moderate budgets, while its benefit eventually saturates as additional computation is allocated.




\section{Conclusion}

We present $\tau_0$-VLA, a hierarchical VLA system for long-horizon manipulation that combines a memory-augmented high-level policy with a low-level policy. The high-level policy handles routine decisions efficiently and allocates world-model-guided computation when a decision benefits from explicit consequence evaluation. The low-level policy is trained from heterogeneous robot and vision-language data for deployable whole-body mobile manipulation. Together, the two levels provide a practical way to allocate reasoning over long task horizons while retaining stable real-robot execution.

\FloatBarrier



\bibliographystyle{plainnat}
\bibliography{references}

\appendices
\section{Author Contributions}

\begingroup
\setlength{\parskip}{0.45em}

\noindent\textbf{High-Level Policy Training and Research:}
Xiaowei Cai, Jingxiao Chen, Xinchen Li, Yifan Li, Yi Liu, Jianlan Luo,
Junwen Miao, Ruiqi Ni, Buqing Nie, Jiaxu Wang, Dafeng Wei, Pengwei Xie,
Pu Yang, Hangjian Ye, Xiangyu Yue, and Jinyu Zhang.

\noindent\textbf{Low-Level Policy Training and Research:}
Xiaowei Cai, Bingao Chen, Jingxiao Chen, Jingshun Huang, Yi Liu,
Jianlan Luo, Junwen Miao, Dafeng Wei, Dongming Wu, Hangjian Ye,
Jinyu Zhang, and Pengfei Zhou.

\noindent\textbf{Training Infra:}
Peiqi Wang, Sen Wang, and Qinglin Zhang.

\noindent\textbf{Robot Infra:}
Tengyu Hou, Dong Li, Zhongyuan Liu, and Xiaoyan Wang.

\noindent\textbf{Writing and Illustration:}
Xiaowei Cai, Yunuo Cai, Jingxiao Chen, Zhi Chen, Yi Liu, Jianlan Luo, Junwen Miao,
Buqing Nie, Dafeng Wei, Dongming Wu, and Jinyu Zhang.

\noindent\textbf{Data Collection and Deployment:}
Zhi Chen, Siyuan Feng, Han Jiang, Runkun Ju, Shaowei Li, Mingjie Pan, Xinlin Ren,
Jianheng Song, and Yue Zhou.

\noindent\textbf{Data Infra:}
Mingxiang Li, Xueyong Zhao and Yue Zhou.

\noindent\textbf{First Authors:}
Jinyu Zhang and Yi Liu.

\noindent\textbf{Corresponding Author:}
Jianlan Luo.

\par\endgroup

\section{Acknowledgments}

We gratefully acknowledge Hongwen Cai, Haoran Chen, Yebin Chen, Zhibo Cui, Zihao Fan, Tao Gao, Yanan Gong, Minghao Gui, Bin He,
Xinyan Hou, Lei Hua, Shuaishuai Li, Tong Li, Qingxiang Liu,
Xiang L\"u, Zongbao Song, Yuna Sun, Zhe Sun, Zhifeng Tang,
Fuming Tian, Jiahao Wan, Hanying Wang, Hao Wang, Sijing Wang,
Tao Wang, Le Yuan, Guozhu Zhang, Hao Zhang, Tianbao Zhang,
Xinbo Zhi, Mingyan Zhou, and Rongxing Zhu for their valuable
contributions to data creation, annotation, and related data
operations. We also thank Yifei Chen for assistance with video
recording and editing.

\section{Additional Details}

This appendix provides additional details on the robot platforms,
physical-robot evaluation protocol, low-level training data, and high-level
policy data construction.

\subsection{Robot Platform Details}
\label{app:robot-platforms}

\noindent\textbf{AGIBOT G1.}
AGIBOT G1 is a wheeled humanoid with an omnidirectional
four-wheel-steering base, dual 7-DoF arms, and modular end-effectors. Our setup
uses parallel-jaw grippers. Its sensing suite includes head-mounted RGB-D and
fisheye cameras and a wrist-mounted camera on each arm.

\noindent\textbf{ARX AC One.}
ARX AC One is a bimanual platform with dual 6-DoF X5
arms and parallel-jaw grippers. We use wrist-mounted cameras together with a
fixed central camera.

\noindent\textbf{Franka Research 3.}
Our bimanual Franka Research 3 setup comprises two
7-DoF torque-controlled arms with custom 3D-printed grippers and three RGB
cameras. Across all three embodiments, the policy receives multi-view RGB
images and proprioceptive state, and the native robot commands are represented
in the unified 40-dimensional action layout defined in
Section~\ref{sec:action-space}.

\subsection{Unified State and Action Representation}
\label{app:unified-space}

The canonical state vector $\mathbf{s}_t$ and each action vector
$\mathbf{a}_{t+j}$, for $j\in\{0,\ldots,H-1\}$, lie in
$\mathbb{R}^{40}$. The stacked action chunk therefore lies in
$\mathbb{R}^{H\times40}$. State and action vectors use the slot ordering shown
in Table~\ref{tab:unified-space}. State channels store the current values, and
the end-effector and joint action channels use the relative parameterization
defined below. The left and right arm blocks follow each embodiment's native
kinematic joint order. A robot with fewer than eight joints per arm fills the
leading entries and masks the remainder.

\begin{table}[H]
\centering
\caption{\textbf{Canonical 40-D state and action layout.} Dimensions are
one-indexed. For a rotation matrix
$\mathbf{R}=[\mathbf{r}_1,\mathbf{r}_2,\mathbf{r}_3]$, we use
$\operatorname{Rot6D}(\mathbf{R})
=[\mathbf{r}_1^\top,\mathbf{r}_2^\top]^\top$.}
\label{tab:unified-space}
\footnotesize
\setlength{\tabcolsep}{3pt}
\renewcommand{\arraystretch}{1.08}
\begin{tabular}{@{}lcl@{}}
\toprule
Coordinates & Dimensions & State representation \\
\midrule
Left EEF position    & 1--3   & Cartesian position in meters \\
Left EEF orientation & 4--9   & $\operatorname{Rot6D}(\mathbf{R}^L)$ \\
Right EEF position   & 10--12 & Cartesian position in meters \\
Right EEF orientation& 13--18 & $\operatorname{Rot6D}(\mathbf{R}^R)$ \\
Left gripper          & 19     & native opening coordinate \\
Right gripper         & 20     & native opening coordinate \\
Waist                  & 21--22 & two native coordinates \\
Planar base velocity  & 23--24 & two native coordinates \\
Left arm joints       & 25--32 & $q^L_1,\ldots,q^L_8$ in radians \\
Right arm joints      & 33--40 & $q^R_1,\ldots,q^R_8$ in radians \\
\bottomrule
\end{tabular}
\end{table}

\noindent\textbf{Relative action encoding.}
Both end-effector and joint actions are expressed relative to the current
state. End-effector actions use position and rotation deltas from the current
pose, with the position delta expressed in the current end-effector frame.
Joint actions use angular offsets from the current joint configuration.

\noindent\textbf{Masks.}
Separate state and action masks identify the valid dimensions for each
embodiment. They activate the applicable end-effector or joint slots, set
unavailable dimensions to zero, and exclude inactive action dimensions from
the training loss.

\noindent\textbf{Control metadata.}
We serialize $\eta$ together with the language command $c_t$ using the
following prompt. The robot type, control mode, and whole-body fields comprise
$\eta$, while the final field contains $c_t$:

\medskip

\noindent\begingroup%
\setlength{\fboxsep}{5pt}%
\colorbox{metabg}{%
  \parbox{\dimexpr\linewidth-2\fboxsep\relax}{%
    \footnotesize
    \ttfamily
    \raggedright
    \setlength{\baselineskip}{1.05\baselineskip}
    You are controlling a robot.\newline
    Robot type: \textless embodiment\textgreater\newline
    Control mode: \textless eef/eef\_wbc/joint\textgreater\newline
    Whole-body control: \textless enabled/disabled\textgreater\newline
    Task: \textless language command\textgreater
  }%
}
\endgroup

\medskip

\noindent\textbf{Masked flow-matching details.}
Using the notation in Eq.~\eqref{eq:masked-flow-path}, the stacked action chunk
$\mathbf{a}_{t:t+H-1}$ lies in $\mathbb{R}^{H\times d_a}$. Given the complete
noisy chunk and the policy conditioning inputs
$(o_t,\mathbf{s}_t,c_t,\eta,\tau)$, the action expert jointly predicts
$\mathbf{v}_{\theta,t+j}^{\tau}\in\mathbb{R}^{d_a}$ for each action token. The
masked flow-matching objective is
\begin{equation}
    \mathcal{L}_{\mathrm{FM}}
    =
    \mathbb{E}\!\left[
    \frac{1}{H\,\operatorname{tr}(\mathbf{M})}
    \sum_{j=0}^{H-1}
    \left\|
    \mathbf{M}\mathbf{v}_{\theta,t+j}^{\tau}
    -
    \mathbf{u}_{t+j}
    \right\|_2^2
    \right].
\end{equation}
The expectation is over training examples, Gaussian noise, and
$\tau=0.001+0.999x$ with $x\sim\operatorname{Beta}(1.5,1)$.
The implementation caps each normalized per-sample loss at $100$ before batch
averaging. At inference, we initialize the chunk from masked Gaussian noise and
integrate the learned field from $\tau=1$ to $\tau=0$. We apply
the action mask to the current noisy chunk before every velocity-field
evaluation and to the final chunk after integration. The reported policies use
an action horizon of $H=30$ and ten uniform Euler updates.

\subsection{Physical-Robot Evaluation Protocol}
\label{app:physical-eval-protocol}

\noindent\textbf{Success rate.}
We apply the same task-specific terminal conditions and milestone annotations
to every method. SR requires every required annotated subtask and the terminal
condition to be completed. A required subtask completed after one or more failed
autonomous attempts still satisfies the SR criterion, provided that no required
subtask is skipped and no task-specific prohibited action occurs.

\noindent\textbf{Progress.}
For each task, we construct a directed acyclic prerequisite graph
$G=(\mathcal{V},\mathcal{E})$ from its original fine-grained annotations. Each
index $i\in\mathcal{V}$ identifies one required milestone. A directed edge
$(i,j)\in\mathcal{E}$ means that milestone $i$ must reach its required task
state before milestone $j$ is eligible for credit. Let
$\operatorname{Anc}(i)$ denote all direct and indirect prerequisites of
milestone $i$.

For trial $r$, let $e_{r,i}=1$ when every milestone in
$\operatorname{Anc}(i)$ has reached its required task state before milestone
$i$, and let $e_{r,i}=0$ otherwise. The local credit
$\alpha_{r,i}\in\{0,0.5,1\}$ is $1$ for first-attempt completion, $0.5$ for
completion after one or more failed autonomous attempts or for a predefined
task-specific partial-completion state, and $0$ otherwise. The
prerequisite-aware credit is
\begin{equation}
    s_{r,i}=e_{r,i}\alpha_{r,i}.
\end{equation}
A completed retry can therefore receive half credit while still enabling later
milestones. A partial-completion state enables descendants only when it meets
the task-specific prerequisite condition. A skipped required milestone receives
no credit and blocks only its descendants. Milestones on independent branches
remain eligible. For example, failure to add the first topping in Make Milk Tea
does not prevent credit for a correctly added second topping. When an execution
record groups multiple subtasks, we resolve the outcomes at the original
subtask granularity using the accompanying action-level annotations.

For a task with $|\mathcal{V}|$ annotated milestones and $T$ evaluated trials,
progress is
\begin{equation}
    \mathrm{Progress}
    =
    \frac{100}{T|\mathcal{V}|}
    \sum_{r=1}^{T}\sum_{i\in\mathcal{V}}s_{r,i}.
\end{equation}
The graph sizes used in Table~\ref{tab:main} are
$|\mathcal{V}|=25$, $14$, $22$, and $13$ for Clean Room, Prepare Ingredients,
Tomato and Egg Stir Fry, and Make Milk Tea, respectively. Book Organization
uses $|\mathcal{V}|=3$, and Collect Laundry uses $|\mathcal{V}|=5$. The three
Tidy Makeup Table groups use $|\mathcal{V}|=2$, $2$, and $4$ and are evaluated
separately.

\noindent\textbf{Task-specific rules.}
Repeated salt addition in Tomato and Egg Stir Fry is a prohibited action rather
than an autonomous retry. It invalidates SR and assigns $0.5$ progress credit
to the salt-addition subtask. Other autonomously completed subtasks are scored
according to the prerequisite graph.

\noindent\textbf{Trial accounting.}
Every method--task entry in Table~\ref{tab:main} uses ten independently
collected physical-robot trials.

Each physical-robot trial uses a fixed wall-clock limit measured from the start
of autonomous execution. The task-specific limits are summarized in
Table~\ref{tab:physical-eval-timeouts}.

\begin{table}[t]
\centering
\caption{\textbf{Maximum duration of each physical-robot trial.}}
\label{tab:physical-eval-timeouts}
\small
\setlength{\tabcolsep}{5pt}
\renewcommand{\arraystretch}{1.10}
\begin{tabular}{lc}
\toprule
Task & Maximum duration \\
\midrule
Clean Room & 20\,min \\
Prepare Ingredients & 20\,min \\
Tomato and Egg Stir Fry & 20\,min \\
Make Milk Tea & 10\,min \\
Book Organization & 5\,min \\
Collect Laundry & 5\,min \\
Tidy Makeup Table (each group) & 5\,min \\
\bottomrule
\end{tabular}
\end{table}

\noindent\textbf{Reset policy.}
Before every trial, the robot is returned to the task-specific standardized
initial pose, and all task objects, tools, articulated furniture, and
consumables are restored to their designated initial states. Object positions
are either restored to fixed designated locations or, when a task specifies
randomized placement, resampled within the same predefined range for every
method. All state carried over from the preceding trial is cleared.

\noindent\textbf{Termination conditions.}
A trial terminates when the task success conditions are satisfied, the
task-specific time limit is reached, or the scene enters an irrecoverable state.
Only the first case is counted as success. All other termination cases are
counted as failures. For progress, only valid milestones completed before
termination receive credit, following the scoring rule in the main text.

\subsection{Low-Level Data Details}
\label{app:low-level-data}

The low-level training corpus combines approximately $23.4$K hours of
internally collected demonstrations with $16.7$K hours from public robot
datasets. The internal data comprise $21.9$K hours on AGIBOT G1, $585$ hours
on AGIBOT G2, $578$ hours on ARX AC One, and $347$ hours on Franka. The public
data include $9.25$K hours of open-source UMI data together with additional
selected datasets, supplying further embodiments, tasks, and contact-rich
manipulation skills.

We additionally interleave multimodal instruction-following and robot-centric
perception examples with the action data. All sources follow a consistent
sampling and preprocessing pipeline throughout low-level training; the
multimodal samples serve as auxiliary supervision for preserving the
backbone's vision-language capabilities.

\subsection{High-Level Policy Data Construction}
\label{app:hl-data}

Supervision for the high-level policy is generated automatically from task instructions, stage descriptions, executable subtask annotations, segmented demonstrations, and videos. In the data pipeline, we denote these three instruction levels as L1, L2, and L3, respectively. Construction proceeds in three steps. \textbf{(1) Pre-labeling.} Using google/Gemma4-31B-it as the annotator, we generate a \texttt{<think>} field that summarizes scene state, progress, constraints, and failures, together with a \texttt{<memory>} field that compresses completed stages while retaining detail for the active subtask. The input includes the carried memory and previously committed subtask. The target \texttt{<memory>} is aligned with the current observation, while the target \texttt{<subtask>} reuses the executable-subtask annotation for what should be executed next. \textbf{(2) Keyframe extraction.} Following each subtask's temporal range, we extract three synchronized views (\texttt{top\_head}, \texttt{hand\_left}, \texttt{hand\_right}) with \texttt{ffmpeg}. \textbf{(3) VQA assembly.} We combine the synchronized views with the generated fields and executable-subtask annotations to form structured training examples for the high-level policy.

To make the high-level policy robust to deployment-time memory misalignment, we perturb \emph{only} the input memory while reading the corrected target from the demonstration. This procedure derives the five instance families in Table~\ref{tab:hl-instances} without additional annotation. Here $\mathcal{M}_n$ denotes the memory upon entering segment $n$. The \emph{within-subtask} family is the aligned case, and the remaining families target distinct failure modes.

\begin{table*}[t]
\centering
\small
\setlength{\tabcolsep}{6pt}
\caption{High-level instance families, synthesized by perturbing only the input memory while reading the corrected target from the demonstration. $\mathcal{M}_n$ is the memory upon entering segment $n$. A single unified \texttt{<think>}/\texttt{<memory>}/\texttt{<subtask>} format instantiates all of them at zero extra annotation.}
\label{tab:hl-instances}
\begin{tabular}{@{}lllll r@{}}
\toprule
Family & Sampling position & Input $\rightarrow$ target memory & Target subtask & Deployment failure countered & Mix \\
\midrule
\emph{within-subtask} & anywhere in seg.\ $n$          & $\mathcal{M}_n \rightarrow \mathcal{M}_n$                 & seg.\ $n$         & --- (aligned, normal progression)     & 58\% \\
\emph{transition}     & tail of seg.\ $n$              & $\mathcal{M}_n \rightarrow \mathcal{M}_{n+1}$             & seg.\ $n{+}1$     & starting a new subtask after completion & 15\% \\
\emph{catch-up}       & head of seg.\ $n$              & $\mathcal{M}_{n-1} \rightarrow \mathcal{M}_n$             & seg.\ $n$         & memory lag (behind the visual state)  & 10\% \\
\emph{rollback}       & late in seg.\ $n$              & $\mathcal{M}_{n+1\ldots n+3} \rightarrow \mathcal{M}_n$   & retry seg.\ $n$   & memory run-ahead (over-optimistic)    & 12\% \\
\emph{error-think}    & annotated failure frame        & $\mathcal{M}_n \rightarrow$ type-dependent                & recovery step     & unnoticed execution failure           & 5\% \\
\bottomrule
\end{tabular}
\end{table*}

For \emph{error-think}, the \texttt{<think>} field first flags the failure and
then repairs memory according to the failure type. Recoverable failures, such
as an empty grasp, are retried with memory unchanged. Failures that undo prior
progress, such as a dropped object, trigger rollback to the preceding subtask.
Samples marked \texttt{restorable}$=$false are skipped, and \emph{rollback}
instances are capped at $10$--$15\%$ to avoid teaching the high-level policy to
distrust correct memory. We additionally apply six-dimensional visually grounded
augmentation to the L1/L2 task and stage instructions to improve instruction
diversity and zero-shot steerability.

\myparagraph{Quality control.}
Quality control operated in two phases. During prompt development, fixed-seed
stratified human spot-checks on small batches surfaced failure modes, chiefly
cross-task contamination, where the annotator hallucinated action constraints
from an unrelated task (e.g.\ milk-tea constraints appearing under a
fruit-shelving task). All visual prompts embed a verification rule that requires
omitting any attribute that cannot be confirmed from the images (``when in
doubt, leave it out''), and each identified failure mode was hardened into a
programmatic filter applied at scale: a cross-task contamination filter, an
empty-\texttt{<subtask>} detector, and a memory-contamination filter. Once the
prompts and filters were fixed, the full corpus was pre-labeled and filtered
programmatically, without further per-sample human inspection.

The contamination filter builds a per-episode noun vocabulary from the L1 and L3
annotations and rejects a constraint whose operated object lies entirely outside
this vocabulary, or whose slot nouns overlap it by less than $20\%$. Tightening
the criterion from a naive disjointness test (discard rate $0.33\%$, with
residual contamination) through an over-aggressive ratio threshold (discard rate
$2.32\%$, which removed many valid but verbose constraints) to the final
two-rule filter with an expanded stop-word list (discard rate $0.42\%$) left
zero residual contamination on a $3.26$M-sample validation set. Structural
dirty-data filtering additionally removes empty-\texttt{<subtask>} episodes and
the downstream memory corruption they induce, using a three-tier policy by
per-task dirty ratio (tasks above $90\%$ dirty are excluded outright, tasks
between $10\%$ and $90\%$ are filtered at the episode level, and tasks below
$10\%$ have only their empty-\texttt{<subtask>} samples removed). This discards
$11.74\%$ of episodes and retains $40.4$M ($88.26\%$) clean samples. The
held-out evaluation set for the high-level policy is drawn from frames unseen in training.

\subsection{Asynchronous Dual-System Serving}
\label{app:async-serving}

The main text describes the logical dependency between high-level subtask
generation and low-level action generation as a sequential inference step. In
deployment, we pipeline these stages asynchronously because a high-level
decision is far slower than one low-level control period. A background worker
continuously recomputes the next subtask for each
active episode and publishes the generated subtask to a per-episode cache
(refreshed about every $1$\,s). The control loop only reads this cache: each tick
takes the cached subtask as the current language command and returns an
action chunk immediately, without waiting for the high-level policy. A slow or failed high-level decision thus
only delays the next command refresh, while $\pi_\theta$ keeps executing the
current cached subtask at the control rate ($\sim\!30$\,Hz).


\subsection{Book Organization Task Settings}
\label{app:book_task_settings}

The Book Organization task contains four books of different heights and four fixed slots.
At the beginning of each trial, the four books are placed in a shuffled order.
At each step, the robot can use its two arms to exchange the positions of any two books.
The task is completed when all four books are arranged from tallest to shortest.
The high-level policy must therefore predict which pair of books should be exchanged next to make progress toward the target ordering.
In the in-domain setting, the initial arrangement of the four books has appeared in the training data.
In the out-of-domain (OOD) setting, the initial arrangement of the four books has not appeared in the training data.
The OOD setting is designed to evaluate the robustness of the high-level policy to observations arising from initial book arrangements not encountered during training.

\subsection{High-Level Evaluation Protocol}
\label{app:high-level-eval-protocol}

The high-level evaluation isolates decision making from physical execution
using annotated samples collected at subtask boundaries. Each sample contains
the task instruction, current robot observation, execution memory when
applicable, the previously committed subtask, and the ground-truth next
subtask. Within each comparison, all methods receive the same samples and
produce the same next-subtask output format. The memory ablation changes only
whether execution memory is provided.
The comparison among Plan Once, Best-of-$N$, and TTC instead holds the
evaluation inputs and output interface fixed while varying the decision-time
inference procedure.

\subsection{LLM-as-a-Judge Protocol}
\label{app:llm-judge}

For the open-loop TTC evaluation, each sample contains a robot observation from a particular stage of a task and the corresponding ground-truth next subtask.
Each method predicts the next subtask from the same input, and an LLM judge evaluates whether the prediction is correct with respect to the ground truth.

Exact string matching can substantially underestimate performance because the same executable subtask may be expressed with different wording. We therefore use GPT-5.4 as a semantic judge. For each sample, the judge is given the task goal, the ground-truth next subtask, and the predicted next subtask, and assigns one of three labels: \textit{equivalent}, \textit{adjacent}, or \textit{wrong}.

A prediction is \textit{equivalent} only if it describes the same immediate physical state transition as the ground truth. The required arm, action, object, destination, and relevant material state must agree, while synonyms and harmless differences in wording or granularity are allowed. A prediction is \textit{adjacent} if it is a physically valid preceding, following, or reorderable step toward the same task goal, but does not match the immediate transition specified by the ground truth. A prediction is \textit{wrong} if it uses an incompatible arm, action, object, destination, or state, hallucinates an action, or incorrectly terminates or resets the task. We do not allow the judge to infer omitted objects, destinations, or second-arm actions.

Only \textit{equivalent} predictions are counted as successful; \textit{adjacent} is retained as a diagnostic category. Each unique (goal, ground truth, prediction) tuple is evaluated by two separate temperature-zero calls. Disagreements receive a third judgment; if all three labels differ, a fourth judgment is obtained. The final label is determined by majority vote.

prompt
You are a strict evaluator of the immediate next executable subtask in a robot plan. Given GOAL, GROUND-TRUTH next subtask, and MODEL-PREDICTED subtask, output one label.

equivalent: the same immediate physical state transition. Required arm(s), action, object(s), destination, and material state must match. Synonyms, capitalization, and harmless wording or granularity differences are allowed.

adjacent: a physically valid preceding, following, or reorderable step for the same goal, but it is a different immediate state transition from the ground truth.

wrong: incompatible arm, action, object, destination, or state; hallucinated/nonsensical action; or done/reset when the ground truth is another action.

Do not infer missing material objects, destinations, or a missing second-arm action. Respond only as compact JSON:{"label":"equivalent|adjacent|wrong","reason":"<=16 words"}.

GOAL: \{task\_goal\}
GROUND-TRUTH next subtask: \{ground\_truth\}
MODEL-PREDICTED subtask: \{prediction\}
Classify.

\subsection{Adaptive Routing Details}
\label{app:adaptive-routing}

At deployment, the high-level policy and low-level policy operate in the closed
loop summarized in Algorithm~\ref{alg:system-inference}. At inference step
$t$, the proposal model produces $z_t^{\mathrm{dir}}$ and $\mathcal{M}_t$ from
$h_t$. We compute the routing decision from token logits already produced by
this forward pass, so routing requires no additional model invocation.

Let $p_i$ be the probability assigned to the generated token at position $i$.
Let $\lambda_i^{(1)}$ and $\lambda_i^{(2)}$ be the largest and second-largest
logits at that position, and define their margin as
\begin{equation}
m_i=\lambda_i^{(1)}-\lambda_i^{(2)}.
\end{equation}
The router uses the mean generated-token probability and the mean logit margin
within the \texttt{<memory>} field:
\begin{equation}
\begin{aligned}
u_t^{\mathrm{all}}
    &= \frac{1}{|\mathcal{I}_{\mathrm{all}}|}
       \sum_{i\in\mathcal{I}_{\mathrm{all}}}p_i, \\
u_t^{\mathrm{mem}}
    &= \frac{1}{|\mathcal{I}_{\mathrm{mem}}|}
       \sum_{i\in\mathcal{I}_{\mathrm{mem}}}m_i,
\end{aligned}
\end{equation}
where $\mathcal{I}_{\mathrm{all}}$ indexes all generated tokens and
$\mathcal{I}_{\mathrm{mem}}$ indexes the tokens in the
\texttt{<memory>} field. The binary routing decision is
\begin{equation}
g_t=
\mathbf{1}\!\left[
u_t^{\mathrm{all}}\leq\delta_{\mathrm{all}}
\lor
u_t^{\mathrm{mem}}\leq\delta_{\mathrm{mem}}
\right],
\end{equation}
where $\mathbf{1}[\cdot]$ is the indicator function and
$\delta_{\mathrm{all}}$ and $\delta_{\mathrm{mem}}$ are routing thresholds.
The statistics and routing rule are shared across tasks. The thresholds are
calibrated separately for each task on held-out validation data.

When $g_t=0$, the system sends $z_t^{\mathrm{dir}}$ directly to the low-level
policy. When $g_t=1$, it searches over candidate subtasks, predicts and scores
their visual outcomes, and conditions the reflective model on the retained
branches to generate $z_t^\star$.

\end{document}